\documentclass{article}

\usepackage[preprint]{neurips_2022}

\usepackage[utf8]{inputenc} % allow utf-8 input
\usepackage[T1]{fontenc}    % use 8-bit T1 fonts
\usepackage{hyperref}       % hyperlinks
\usepackage{url}            % simple URL typesetting
\usepackage{booktabs}       % professional-quality tables
\usepackage{amsfonts}       % blackboard math symbols
\usepackage{nicefrac}       % compact symbols for 1/2, etc.
\usepackage{microtype}      % microtypography
\usepackage[dvipsnames]{xcolor}         % colors

\usepackage{amsmath}
\usepackage{amssymb}
\usepackage{mathtools}
\usepackage{amsthm}
\usepackage{graphicx}
\usepackage{hyperref}[colorlinks=true, allcolors=blue]
\usepackage{subcaption}
\newsavebox{\bigimage}
\DeclarePairedDelimiter\ceil{\lceil}{\rceil}

\usepackage{multirow}
\usepackage{hhline}
\usepackage{color, colortbl}
\definecolor{Gray}{gray}{0.9}
\usepackage{rotating}
\usepackage{tablefootnote}
\usepackage{algorithm}
\usepackage{algorithmic}

\title{An Empirical Evaluation of Posterior Sampling for Constrained Reinforcement Learning}

\author{%
  Danil Provodin$^{1,2}$, Pratik Gajane$^1$, Mykola Pechenizkiy$^{1,4}$, Maurits Kaptein$^{2,3}$ \\
  $^1$Eindhoven University of Technology, Eindhoven, The Netherlands\\
  $^2$Jheronimus Academy of Data Science, ‘s-Hertogenbosch, The Netherlands \\
  $^3$Tilburg University, Tilburg, The Netherlands \\
  $^4$University of Jyväskylä, Jyväskylä, Finland \\
  \texttt{ \{d.provodin,p.gajane,m.pechenizkiy\}@tue.nl, M.C.Kaptein@tilburguniversity.edu} \\
}

\begin{document}

\maketitle

\begin{abstract}
  We study a posterior sampling approach to efficient exploration in constrained reinforcement learning. Alternatively to existing algorithms, we propose two simple algorithms that are more efficient statistically, simpler to implement and computationally cheaper. The first algorithm is based on a linear formulation of CMDP, and the second algorithm leverages the saddle-point formulation of CMDP. Our empirical results demonstrate that, despite its simplicity, posterior sampling achieves state-of-the-art performance and, in some cases, significantly outperforms optimistic algorithms. 
\end{abstract}

\section{Introduction}
Reinforcement learning (RL) problem is a sequential decision-making paradigm that aims to improve an agent’s behavior over time by interacting with the environment. In standard reinforcement learning, the agent learns by trial and error based on the scalar signal, called the reward, it receives from the environment, aiming to maximize the average reward. Nonetheless, in many settings this is insufficient, because the desired properties of the agent behavior are better described using constraints. For example, a robot should not only fulfill its task, but should also control its wear and tear by limiting the torque exerted on its motors \citep{tessler2018reward}; recommender platforms should not only focus on revenue growth but also optimize users long-term engagement \citep{asfar2021_RLforRS:survey}. In this paper we study constrained reinforcement learning under the infinite-horizon average reward setting, which encompasses many real-world problems.

A standard way of formulating the constrained RL problem is by the agent interacting with a constrained Markov Decision Process (CMDP) \citep{Altman99constrainedmarkov}, where the agent must satisfy constraints on expectations of auxiliary costs. There are several approaches in the literature that have focused on solving CMDPs. These methods are mainly based on model-free RL algorithms \citep{Chow_2017, Achiam_CMDP2017, tessler2018reward, Bohez2019ValueCM} and model-based RL algorithms \citep{Efroni_2020_CMDP, NEURIPS2020_Brantley, Singh_CMDP_2020}. While model-free methods may eventually converge to the true policy, they are notoriously sample inefficient and have no theoretical guarantees. These methods have demonstrated prominent successes in artificial environments but extend poorly to real-life scenarios. Model-based algorithms, in its turn, focus on sample efficient exploration. From a methodological perspective, most of these methods leverage \textit{optimism in the face of uncertainty} (OFU) principle to encourage adaptive exploration.

We study an alternative approach to efficient exploration in constrained reinforcement learning, \textit{posterior sampling}. Our work is motivated by several advantages of posterior sampling relative to optimistic algorithms described in \cite{Osband_PSRL2013}. First, since posterior sampling only requires solving for an optimal policy for a single sampled CMDP, it is computationally efficient relative to many optimistic methods, which require simultaneous optimization across a family of plausible environments \citep{Efroni_2020_CMDP, Singh_CMDP_2020}. Second, optimistic algorithms require explicit construction of the confidence bounds based on observed data, which is a complicated statistical problem even for simple models. By contrast, in posterior sampling, uncertainty about each policy is quantified in a statistically efficient way through the posterior distribution, which is simpler to implement. Finally, the presence of an explicit prior allows an agent to incorporate known environment structure naturally. This is crucial for most practical applications, as learning without prior knowledge requires exhaustive experimentation in each possible state.

As such, we propose two simple algorithms based on the posterior sampling for constrained reinforcement learning in the infinite-horizon setting:

\begin{itemize}
    \item \textbf{Posterior Sampling of Transitions} maintains posteriors for the transition function while keeping rewards and costs as an empirical mean. At each episode, it samples an extended CMDP from the posterior distribution and executes the optimal policy. To solve the planning problem, we introduce a linear program (LP) in the space of occupancy measures. 
    
    \item \textbf{Primal-Dual Posterior Sampling} exploits a primal-dual algorithm to solve the saddle-point problem associated with a CMDP. It performs incremental updates both on the primal and dual variables. This reduces the computational cost by using a simple dynamic programming approach to compute the optimal policy instead of solving a constrained optimization problem.
\end{itemize}

We provide a comprehensive comparison of OFU- and posterior sampling-based algorithms across three environments in our study. In all cases, despite its simplicity, posterior sampling achieves state-of-the-art results and, in some cases, significantly outperforms other alternatives. In addition, as we show in Section \ref{subsec:results}, posterior sampling is naturally suited to more complex settings where design of an efficiently optimistic algorithm might not be possible.

\section{Related work}
\label{lit_review}

Recently, online learning under constraints has received extensive attention. Over the past several years, learning in CMDPs has been heavily studied in different settings -- episodic, infinite-horizon discounted reward, and infinite-horizon average reward. 
%While our work focuses on the infinite-horizon average reward setting (which imposes exceptional difficulties), other settings have played an essential role in investigating and understanding the infinite-horizon average reward setting. 
Below we provide a short overview of the most relevant works to ours, which are summarized in Table \ref{tbl:lit_rev}.

\paragraph{OFU-based algorithms.} Several OFU-based algorithms have been proposed for learning policies for CMDPs. All the algorithms in this group extend the idea of the UCRL2 algorithm \cite{JMLR:v11:jaksch10a} (the first algorithm based on the optimism principle applied for classical reinforcement learning), but each of them utilizes optimistic exploration in different forms. 

\cite{Efroni_2020_CMDP} and \cite{NEURIPS2020_Brantley} consider sample efficient exploration in finite-horizon setting via \textit{double optimism}: overestimation of rewards and underestimation of costs. This approach makes each state-action pair more appealing along both rewards and costs simultaneously. Conversely, \cite{pmlr-v120-zheng20a} and \cite{Liu_2021} considers \textit{conservative (safe) exploration} in infinite and finite horizon settings correspondingly using optimism (overestimation) over rewards and pessimism (overestimation) over costs. Unlike previous algorithms, where the reward signal is perturbed to one side or the other, \cite{Singh_CMDP_2020} considers \textit{optimism over transition probabilities} in an infinite-horizon setting. Specifically, the authors construct a confidence set of transition probabilities and choose "optimistic" empirical transitions to force the exploration at each step. Consequently, \cite{Efroni_2020_CMDP}, \cite{qiu_2021_cmdp_posterior}, and \cite{Liu_2021} leverage the saddle-point formulation of CMDP in the episodic setting. They use standard dynamic programming for the policy update of the primal variable with the scalarized reward function. Recently, \cite{chen_2022_optimisQlearn} considered \textit{optimistic Q-learning} in infinite-horizon CMDP providing a tighter bound on the span of the bias function and strictly improves the result of \cite{Singh_CMDP_2020}.

\paragraph{Posterior sampling algorithms.} Somewhat surprisingly, posterior sampling has not yet gained such traction in constrained reinforcement learning. Although it has been extensively researched in unconstrained setting \citep{Osband_PSRL2013, AY_bayesian_control_2015, NIPS2017_Shipra_OPSRL, TS_MDP_Ouyang_2017}, we are aware of only one work on posterior sampling in CMDPs \citep{Agarwal_2021_PSRL}. That article provides a theoretical analysis of posterior sampling for CMDP with known reward and cost functions in the infinite-horizon average reward setting. Specifically, their algorithm samples transitions from a posterior distribution at each episode and solves an LP problem to find the optimal policy. Our Posterior Sampling of Transitions algorithm is an extension of the algorithm from \cite{Agarwal_2021_PSRL} to unknown reward and cost functions.

\newcolumntype{g}{>{\columncolor{Gray}}c}
\begin{table}[htbp]
  \centering
  \begin{tabular}{lggggggg}
    \hline
    \rowcolor{white}
    \multicolumn{2}{c}{Algorithm} & \multicolumn{2}{c}{Regret} & \multicolumn{2}{c}{Constraint violation} & \multicolumn{2}{c}{Exploration} \\
    
    \hline
    \rowcolor{white}
    &&&&&&& \\
    \rowcolor{white}
    & \multirow{-2}{*}{\cite{Efroni_2020_CMDP}} && \multirow{-2}{*}{$\tilde{O}(H^3 \sqrt{S^2AK})$} && \multirow{-2}{*}{$\tilde{O}(H^3 \sqrt{S^2AK})$} && \multirow{-2}{*}{$r,c$}  \\
    \cline{2-8}
    
    &  &&&&&& \\
    & \multirow{-2}{*}{\cite{NEURIPS2020_Brantley} 
    %\tablefootnote{Although this algorithm is initially developed for episodic tasks, we borrow its principle with slight adjustment and apply it for infinite-horizon setting.}
    } && \multirow{-2}{*}{$\tilde{O}(H^3 \sqrt{S^3A^2K})$} && \multirow{-2}{*}{$\tilde{O}(H^3 \sqrt{S^3A^2K})$} &&  \multirow{-2}{*}{$r,c$} \\
    \cline{2-8}
   
    \rowcolor{white}
    &&&&&&& \\
    \rowcolor{white}
    & \multirow{-2}{*}{\cite{qiu_2021_cmdp_posterior}} && \multirow{-2}{*}{$\tilde{O}(H \sqrt{S^2AK})$}  && \multirow{-2}{*}{$\tilde{O}(H \sqrt{S^2AK})$} && \multirow{-2}{*}{$p$}  \\
    \cline{2-8}
    
    \rowcolor{white}
    \multirow{-7}{*}{\rotatebox{90}{\parbox[c]{3cm}{\centering finite horizon}}} &&&&&&& \\
    \rowcolor{white}
    & \multirow{-2}{*}{\cite{Liu_2021}} && \multirow{-2}{*}{$\tilde{O}(H^3 \sqrt{S^3AK})$}  && \multirow{-2}{*}{$\tilde{O}(1)$} && \multirow{-2}{*}{$r,c$}  \\

    \hline
    & %\multirow{2}{*}{\cite{NEURIPS2019_cheung_2019}} &&    \\   
    %&&&&& & \\\cline{2-8}
     &&&&&& \\
    & 
    \multirow{-2}{*}{\cite{Singh_CMDP_2020}} && \multirow{-2}{*}{$\tilde{O}(\sqrt{SA}T^{2/3} )$} && \multirow{-2}{*}{$\tilde{O}(\sqrt{SA}T^{2/3} )$} && \multirow{-2}{*}{$p$}  \\
   \cline{2-8}
   
   &  &&&&&& \\
     & 
     \multirow{-2}{*}{\cite{pmlr-v120-zheng20a}} &&  \multirow{-2}{*}{$\tilde{O}(mSAT^{3/4} )$} && \multirow{-2}{*}{0} && \multirow{-2}{*}{$r,c$} \\
   \cline{2-8}
   
    \rowcolor{white}
    &&&&&&& \\
    \rowcolor{white}
    & \multirow{-2}{*}{\cite{chen_2022_optimisQlearn} \tablefootnote{This algorithms uses unconventional terms which complicates the comparison based on other parameters}} && \multirow{-2}{*}{$\tilde{O}(\sqrt{T} )$} && \multirow{-2}{*}{$\tilde{O}(\sqrt{T} )$} && \multirow{-2}{*}{$Q$-func.}  \\
   
   \hhline{~=======}
    \multirow{-7}{*}{\rotatebox{90}{\parbox[c]{3cm}{\centering infinite horizon}}} &  &&&&&& \\
    & 
    \multirow{-2}{*}{\cite{Agarwal_2021_PSRL}} &&  \multirow{-2}{*}{$\tilde{O}( poly(DSA) \sqrt{T})$} && \multirow{-2}{*}{$\tilde{O}( poly(DSA) \sqrt{T})$} && \multirow{-2}{*}{$p$} \\
   \hline
  \end{tabular}
  \caption{Summary of work on provably efficient constrained reinforcement learning. $S$ and $A$ represent number of states and actions; and $m$ is the number of constraints. $H$ is the maximum length of an episode and $K$ is the number of episodes in finite horizon setting. $T$ is the total horizon in infinite horizon setting and, finally, $D$ is the diameter of CMDP. $r,c$, and $p$ are elements of CMDP -- rewards, costs, and transitions correspondingly. The "Exploration" column shows the source of the exploration in the algorithm, roughly speaking, these are elements of CMDP where a bonus term is being added (except for \cite{Agarwal_2021_PSRL}, which considers posterior sampling over $p$). Algorithms that are further used in the empirical comparison are highlighted in gray.}
  \label{tbl:lit_rev}
\end{table}

\section{Problem formulation}
\label{sec:problem_formulation}

\paragraph{CMDP.} A constrained MDP model is defined as a tuple $\mathcal{M} = (\mathcal{S}, \mathcal{A}, p, \textit{\textbf{r}}, \tau, \rho)$ where $\mathcal{S}$ is the state space, $\mathcal{A}$ is the action space, $p : \mathcal{S} \times \mathcal{A} \xrightarrow{} \Delta_{\mathcal{S}}$ is the unknown transition function, where $\Delta_{\mathcal{S}}$ is simplex over $\mathcal{S}$, $\textit{\textbf{r}} : \mathcal{S} \times \mathcal{A} \xrightarrow{} [0, 1]^{m+1}$ is the unknown reward vector function of interest, $\tau$ is a cost threshold, and $\rho$ is the known initial distribution of the state. In general, CMDP is an MDP with multiple reward functions ($r_0, r_1,\dots,r_m$), one of which, $r_0$, is used to set the optimization objective, while the others, ($r_1,\dots,r_m$), are used to restrict what policies can do.

For any policy $\pi$ and initial distribution $\rho$, the expected infinite-horizon average reward is defined as

\begin{equation}
    J^{\pi}(r, \rho) = \overline{\lim}_{T \to \infty} \frac{1}{T} \sum_{t=0}^T \mathbb{E}_{\rho}^{\pi} \left [ r(S_t,A_t) \right ]
\end{equation}
where $\mathbb{E}_{\rho}^{\pi}$ is the expectation under the probability measure $\mathbb{P}_{\rho}^{\pi}$ over the set of infinitely long state-action trajectories, $\mathbb{P}_{\rho}^{\pi}$ is induced by policy $\pi$ (which directs what actions to take given what history), and the initial state $s \sim \rho$. Given some fixed initial state distribution  $\rho$  and reals  $\tau_1, \dots , \tau_m \in \mathbb{R}$ , the CMDP optimization problem is to find a policy  $\pi$  that maximizes $J^\pi(r_0,\rho)$ subject to the constraints  $J^\pi(r_i,\rho) > \tau_i, i = 1, \dots, m$:

\begin{equation}
    \label{eq:objective}
    \max_\pi J^\pi(r_0,\rho) \quad \text{ s.t. } \quad  J^\pi(r_i,\rho) > \tau_i,\, i=1,\dots,m \,. % \tag{$\text{CMDP}_\mu$}
\end{equation}

Sometimes it is convenient to define CMDP through main scalar reward function $r$ and $m$ cost functions ($c_1, \dots, c_m$). In order to do that, one can easily recast the definition of CMDP by multiplying original reward components ($r_1, \dots, r_m$) by $-1$ and , with slight abuse of notation, denoting ($r_0, r_1,\dots,r_m$) as ($r, c_1, \dots, c_m$). Then, in terms of costs, CMDP $\mathcal{M} = (\mathcal{S}, \mathcal{A}, p, r, c, \tau, \rho)$, and optimization problem \eqref{eq:objective} becomes

\begin{equation}
    \max_\pi J^\pi(r,\rho) \quad \text{ s.t. } \quad  J^\pi(c_i,\rho) \leq \tau_i,\, i=1,\dots,m \,. % \tag{$\text{CMDP}_\mu$}
    \label{eq:objective_cost}
\end{equation}

In contrast to episodic RL problems, in which the state is reset at the beginning of each episode, infinite-horizon RL problems appear to be much more challenging as the interaction between agent and environment never ends or resets. Consequently, in order to control the regret vector (defined below), we assume that the CMDP $\mathcal{M}$ is unichain, i.e., for each stationary policy $\pi$, the Markov chain induced by $\pi$  contains a single recurrent class, and possibly, some transient states. Where a stationary policy $\pi$, in its turn, is a mapping from state space $\mathcal{S}$ to a probability distribution on the action space $\mathcal{A}$, $\pi : \mathcal{S} \xrightarrow{} \Delta_{\mathcal{A}}$, which is independent of $t$, i.e., does not change over time.

\paragraph{Linear Programming approach for solving CMDP.} When CMDP is known and unichain, an optimal policy for \eqref{eq:objective_cost} can be obtained by solving the following liner program (LP) \citep{Altman99constrainedmarkov}:

\begin{align}
    \max_{\mu} \sum_{s,a} \mu(s,a) r(s,a), \label{eq1}\\
    \mathrm{s.t.}\quad \sum_{s,a} \mu(s,a) c_i(s,a) \leq \tau_i,\, \quad i=1,\dots,m, \\
    \sum_a \mu(s,a) = \sum_{s', a} \mu(s', a) p(s',a,s), \quad \forall s \in \mathcal{S}, \\
    \mu(s,a) \geq 0, \quad \forall (s,a) \in \mathcal{S} \times \mathcal{A}, \quad \sum_{s,a} \mu(s,a) = 1, \label{eq4}
\end{align}
where the decision variable $\mu(s,a)$ is occupancy measure (fraction of visits to $(s,a)$). 

Given the optimal solution for LP \eqref{eq1}-\eqref{eq4}, $\mu^*(s,a)$, one can construct the optimal stationary policy $\pi^*(a|s)$ for \eqref{eq:objective_cost} by choosing action $a$ in state $s$ with probability $\frac{\mu^*(s,a)}{\sum_{a'} \mu^*(s,a')}$.

\paragraph{Regret vector.} In order to measure the performance of a learning algorithm we define its reward and cost regret. The cumulative reward and cost regret for the $i$-th cost until time $T$ is defined as

\begin{equation}
    Reg_+ (T;r) = \sum_{t=1}^T \left [ r^* - r(s_t, a_t) \right ]_+ \quad \text{and} \quad Reg_+ (T;c_i) = \sum_{t=1}^T \left [ c_i(s_t, a_t) - \tau_i \right ]_+,
\end{equation}
respectively. Above $[x]_+ := \max \{0,x\}$ and $r^*$ is the optimal average reward of the CMDP \eqref{eq:objective_cost}. In fact, $r^*$ is also the optimal value of LP \eqref{eq1}-\eqref{eq4}.  Note that in constrained setting, the immediate reward regret, $ r^* - r(s_t, a_t)$, might be negative since a policy violate the constraints.

\section{Posterior sampling algorithms}
\label{posterior_algos}

In this section, we introduce two simple algorithms based on posterior sampling. Both algorithms use the doubling epoch framework of \citet{JMLR:v11:jaksch10a}. The rounds $t= 1,...,T$ are broken into consecutive epochs as follows: the $k$-th epoch begins at the round $t_k$ immediately after the end of $(k - 1)$-th epoch and ends at the first round $t$ such that  $N_{t}(s,a) \geq 2 N_{t_k}(s,a)$ for some state-action pair $(s,a)$, i.e., within epoch $k$ an algorithm has visited some state-action pair $(s,a)$ at least the same number of times it had visited this pair $(s,a)$ before epoch $k$ started. An algorithm computes a new policy $\pi_k$ at the beginning of every epoch $k$, and uses that policy through all the rounds in that epoch. The policy $\pi_k$ to be used in epoch $k$ is computed as the optimal policy of an extended CMDP $\hat{\mathcal{M}}_k = (\mathcal{S}, \mathcal{A}, p_{t_k}, \Bar{r}_{t_k}, \Bar{c}_{t_k}, \tau, \rho)$ defined by the sampled transition probability vectors ($p_{t_k}$) and empirical reward and cost functions defined as:

\begin{equation}
    \label{mean_rewards}
    \Bar{r}_t(s,a) = \frac{\sum_{j=1}^{t-1} \mathbb{I} \{ s_t=s, a_t=a \} r_t}{N_t(s,a) \lor 1}, \quad \forall s \in \mathcal{S}, a \in \mathcal{A},
\end{equation}

\begin{equation}
    \label{mean_costs}
    \Bar{c}_{i,t}(s,a) = \frac{\sum_{j=1}^{t-1} \mathbb{I} \{ s_t=s, a_t=a \} c_{i,t}}{N_t(s,a) \lor 1}, \quad \forall s \in \mathcal{S}, a \in \mathcal{A}, \quad i = 1 \dots, m,
\end{equation}
where $N_t(s,a)$ and $N_t(s,a,s')$ denote the number of visits to $(s,a)$ and $(s,a,s')$ respectively.

We present a posterior sampling algorithms using Dirichlet priors. Dirichlet distribution is a convenient choice maintaining posteriors for the transition probability vectors $p(s,a)$, as they satisfy the following useful property: given a prior $Dir(\alpha_1,...,\alpha_S)$ on $p(s,a)$, after observing a transition from state $s$ to $i$ (with underlying probability $p_i(s,a)$), the posterior distribution is given by $Dir(\alpha_1,...,\alpha_i+ 1,...,\alpha_S)$. By this property, for any $(s,a) \in \mathcal{S} \times \mathcal{A}$, on starting from prior $Dir(\textbf{1})$ for $p(s,a)$, where $\textbf{1}$ is vector of ones, the posterior at time $t$ is $Dir\left ( \{N_t(s,a,s')\}_{s' \in \mathcal{S} } \right )$.

\paragraph{Posterior Sampling of Transitions.}
The first algorithm we propose proceeds similarly to many optimistic algorithms, i.e., at the beginning of every epoch $k$, posterior sampling of transitions algorithm solves LP \eqref{eq1}-\eqref{eq4} substituting unknown parameters $p, r, c_1, ..., c_m$ with the sampled transition vectors $p_{t_k} \sim Dir \left ( \{ N_{t_k}(s,a,s') \}_{s' \in \mathcal{S}} \right )$ and empirical estimates $\Bar{r}_{t_k}, \Bar{c}_{1,t_k}, ..., \Bar{c}_{m,t_k}$ defined in \eqref{mean_rewards}, \eqref{mean_costs}. The  algorithm is summarized as Algorithm \ref{alg1:psrl_transitions}. 

\begin{algorithm}[tb]
   \caption{Posterior Sampling of Transitions (\textit{PSRLTransitions})}
   \label{alg1:psrl_transitions}
\begin{algorithmic}[1]
    \STATE {\bfseries Input:} -

    \STATE Initialization: $t \gets 1$, $t_k \gets 1$
    \FOR{episodes $k = 1, 2, \dots$}
        \STATE $t_k \gets t$
        \STATE Update $\Bar{r}_k(s,a)$ and $\Bar{c}_{i,k}(s,a)$ for $i=1, \dots, m$ as in (\ref{mean_rewards}) and (\ref{mean_costs})
        \STATE Generate $p_k(\cdot|s,a) \sim Dir \left ( \{ N_{t_k}(s,a,s') \}_{s' \in \mathcal{S}} \right )$
        \STATE Compute $\pi_k(\cdot)$ by solving LP (\ref{eq1})-(\ref{eq4}) with  $p_k(\cdot|s,a), \Bar{r}_k(s,a)$ and $\Bar{c}_{i,k}(s,a)$
        \REPEAT
            \STATE $a_t = \pi_k(s_t)$
            \STATE Observe new state $s_{t+1}$
            \STATE Update counters $N_t(s_t,a_t)$ and $N_t(s_t,a_t,s_{t+1})$
            \STATE $t \gets t + 1$
        \UNTIL{$N_t(s,a) \leq 2 N_{t_k}(s,a)$} for some $(s,a) \in \mathcal{S} \times \mathcal{A}$
    \ENDFOR

\end{algorithmic}
\end{algorithm}

The algorithm has to solve an LP problem with $O(SA)$ constraints and decision variables $K_T$ times, where $K_T$ is number of epochs during horizon $T$. \footnote{$K_T \leq \sqrt{2SAT \log T}$, see, e.g., \cite{TS_MDP_Ouyang_2017}} Although it is computationally more efficient than algorithms that require simultaneous optimization across a family of plausible environments, in the limit of large state or action spaces, solving such linear program can become a formidable computational burden. For that reason, we propose a primal-dual posterior sampling algorithm below.

\paragraph{Primal-Dual Posterior Sampling.}
To overcome the limitation above, we consider a heuristic for the Algorithm \ref{alg1:psrl_transitions}. Specifically, we consider the Lagrangian relaxation of problem (\ref{eq:objective_cost}):

\begin{equation}
    \min_{\lambda: \lambda(i) \geq 0, \forall i} \max_{\pi} J^{\pi}(r, \rho) + \lambda^T \left ( \tau -  J^{\pi}(c, \rho) \right ),
    %L(\pi, \lambda) = J^{\pi}(r, \rho) + \lambda^T \left ( \tau -  J^{\pi}(c, \rho) \right ),
    \label{lagr_relax}
\end{equation}
where $\tau$ and $J^{\pi}(c, \rho)$ are vectors from $\mathbb{R}^m$ composed of $\tau_i$ and $J^{\pi}(c_i, \rho)$ respectively for $i = 1, \dots, m$. For a fixed choice of Lagrange multipliers $\{ \lambda(i) \}_{i=1}^m$, program \eqref{lagr_relax} is an unconstrained optimization problem with a pseudo-reward $r_{\lambda}(s,a)$ defined as:

\begin{equation*}
    r_{\lambda}(s, a) = r(s,a) + \sum_{i=1}^m \lambda(i) \left ( \tau_i -  c_i(s, a) \right )
\end{equation*}
Thus, Lagrangian relaxation squeezes the original CMDP to a standard MDP (i.e., process with the same transitions and the modified, but scalar, reward function). 

From MDP theory \citep{Bert12}, we know that if the MDP is unichain (and even weakly communicating), the optimal average reward $J(r, \rho) = \max_{\pi} J^{\pi}(r, \rho)$ satisfies the Bellman equation

\begin{equation}
    J(r, \rho) + v(s) = \max_{a \in \mathcal{A}} \left \{ r(s,a) + \sum_{s' \in \mathcal{S}} p(s'|s,a) v(s') \right \}
    \label{bel_eq}
\end{equation}
for all $s \in \mathcal{S}$. This allows to reduce the computational cost by performing a policy update using  standard dynamic programming instead of solving LP \eqref{eq1}-\eqref{eq4}. 

Algorithm \ref{alg2:primal_dual_psrl} formalized the logic above and presents the Primal-Dual Posterior Sampling algorithm. The algorithm consists of two main steps: \textit{Policy update} and \textit{Dual update}. For policy update we use a value iteration method of \citep{Bertsekas_VI} and for dual update we use a projected gradient descent \citep{Zinkevich_ogd_2003}. It is worth mentioning that the reduced computational cost comes with a flaw: by construction of value iteration algorithm, policy $\pi_k$ is deterministic as the optimal solution for an extended CMDP $\hat{\mathcal{M}}_k$ with scalarized reward function $r_{\lambda_k}$. However, such a policy cannot be optimal in the original CMDP $\hat{\mathcal{M}}_k$ if in there all deterministic policies are suboptimal. To counter this flaw, we execute the mixture policy $\tilde{\pi}_k$ instead of $\pi_k$, defined in line 10 of Algorithm \ref{alg2:primal_dual_psrl}.

%Equation \eqref{lagr_relax} can be considered as a game between two players: the agent $\pi$ and the Lagrange multiplier $\lambda$. This process is expected to converge to the Nash equilibrium with value

%\begin{equation}
%       L^* = \min_{\pi} \max_{\lambda}  L(\pi, \lambda).
%\end{equation}
%Furthermore, strong duality is known to hold for CMDP \citep{Altman99constrainedmarkov} and thus the expected value of this game is expected to converge to $L^* = r^*$.

\begin{algorithm}[tb]
   \caption{Primal-Dual Posterior Sampling (\textit{PSRLLagrangian})}
   \label{alg2:primal_dual_psrl}
\begin{algorithmic}[1]
    \STATE {\bfseries Input:} Learning rate $\eta$

    \STATE Initialization: $t \gets 1$, $t_k \gets 1$, $\tilde{\pi}_0 = \textbf{0}$
    \FOR{episodes $k = 1, 2, \dots$}
        \STATE $t_k \gets t$
        \STATE Update $\Bar{r}_k(s,a)$ and $\Bar{c}_{i,k}(s,a)$ for $i=1, \dots, m$ as in (\ref{mean_rewards}) and (\ref{mean_costs})
        \STATE Generate $p_k(\cdot|s,a) \sim Dir \left ( \{ N_{t_k}(s,a,s') \}_{s' \in \mathcal{S}} \right )$
        \STATE \textit{(Policy update)} Compute $\pi_k(\cdot)$ from (\ref{bel_eq}), where $r_{\lambda_k}(s,a) = \Bar{r}_k(s,a) + \lambda_k^T \left ( \tau -  \Bar{c}_{k}(s,a) \right )$
        \STATE \textit{(Dual update)} $\lambda_{k+1} = \max \{ 0, \lambda_k + \eta [J^{\pi_k}(\Bar{c}, \rho) - \tau ] \}$
        \REPEAT
            \STATE $a_t = \tilde{\pi}_k(s_t)$, where $\tilde{\pi}_k = \tilde{\pi}_{k-1} + \frac{1}{k} [\pi_k - \tilde{\pi}_{k-1} ]$ -- mixture policy
            \STATE Observe new state $s_{t+1}$
            \STATE Update counters $N_t(s_t,a_t)$ and $N_t(s_t,a_t,s_{t+1})$
            \STATE $t \gets t + 1$
        \UNTIL{$N_t(s,a) \leq 2 N_{t_k}(s,a)$} for some $(s,a) \in \mathcal{S} \times \mathcal{A}$
    \ENDFOR

\end{algorithmic}
\end{algorithm}

\section{Empirical evaluation}

This section presents a comprehensive empirical analysis of the proposed algorithms. Specifically, we compare posterior sampling algorithms with the state-of-the-art OFU-based algorithms (benchmarks) across three environments. Among the algorithms mentioned in Table \ref{tbl:lit_rev}, we choose the most diverse but, at the same time, the most comparable for empirical analysis. 

We use three OFU-based algorithms from the existing literature for comparison. These algorithms are based on the first three types of optimistic exploration mentioned in Section \ref{lit_review}: double optimism, conservative optimism, and optimism over transitions. We compare the posterior sampling algorithms to \textit{ConRL} \citep{NEURIPS2020_Brantley}, \textit{C-UCRL} \citep{pmlr-v120-zheng20a} and \textit{UCRL-CMDP} \citep{Singh_CMDP_2020}. Although \textit{ConRL} was originally developed for the episodic setting, we extend it to the infinite-horizon setting by using the doubling epoch framework described in Section \ref{posterior_algos}. Since the definitive algorithm is quite different, we rename it as \textit{CUCRLOptimistic}. To avoid ambiguities we also rename \textit{C-UCRL} and \textit{UCRL-CMDP} algorithms as \textit{CUCRLConservative} and \textit{CUCRLTransitions} correspondingly. 

Below we present a description of environments used in experiments and demonstrate the empirical results. A more detailed description of benchmarks can be found in Appendix \ref{benchmarks}.

\subsection{Environments}
\label{subsec:envs}

\iffalse
\begin{figure}
    \centering
    \begin{subfigure}{.35\textwidth}
        \includegraphics[width=\textwidth]{pics/envs/gridworld4x4.png}
        \caption[gridworld]{Marsrover 4x4}
        \label{gridworld}
    \end{subfigure}\hfill
    \begin{subfigure}{.35\textwidth}
        \includegraphics[width=\textwidth]{pics/envs/box_main.png}
        \caption[box_main]{Box (Main)}
        \label{box_main}
    \end{subfigure}\hfill
    \begin{minipage}[b]{.251\textwidth}
        \begin{subfigure}{.6\linewidth}
                \includegraphics[width=\textwidth]{pics/envs/box_left.png}
                \caption[box_left]{Box (Safe)}
                \label{box_left}
        \end{subfigure}\\
        \begin{subfigure}{.6\linewidth}
                \includegraphics[width=\textwidth]{pics/envs/box_down.png}
                \caption[box_down]{Box (Fast)}
                \label{box_down}
        \end{subfigure}
    \end{minipage}
    \caption{Gridworld environments. The initial position is light green, the goal is dark green, the walls are gray, and risky states are purple. Figure (\ref{gridworld}) illustrates 4x4 Marsrover environment. Figure (\ref{box_main})-(\ref{box_down}) illustrates Box environment. In both cases, the agent's task is to get from the initial state to the goal state, and the optimal policy combines with some probabilities fast and safe ways, which are indicated by arrows on the pictures.}
    \label{envs}
\end{figure}
\fi

\begin{figure}
  \centering
  \begin{subfigure}[b]{0.399\textwidth}
    \centering
    \includegraphics[width=\textwidth]{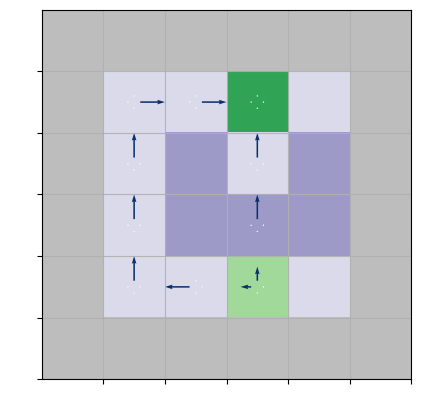}
    \caption[gridworld]{Marsrover 4x4}
    \label{gridworld}
  \end{subfigure}
  \begin{subfigure}[b]{0.399\textwidth}
    \centering
    \includegraphics[width=\textwidth]{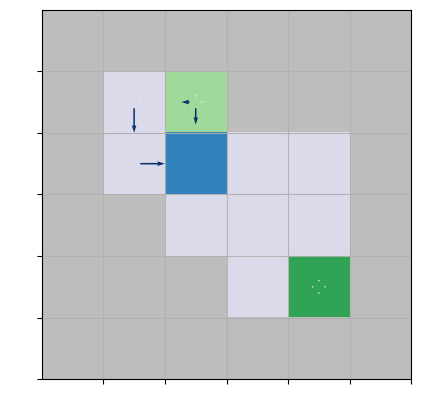}
    \caption[box_main]{Box (Main)}
    \label{box_main}
  \end{subfigure}
  \begin{subfigure}[b]{0.1715\textwidth}
    \centering
    \includegraphics[width=\textwidth]{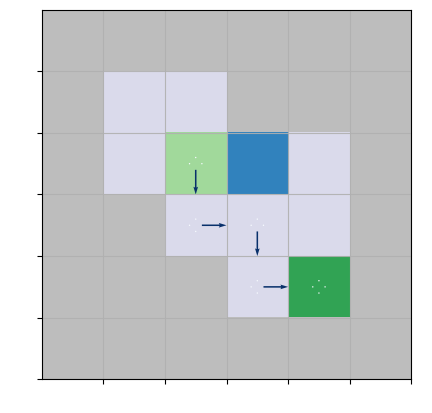}
    \caption[box_left]{Box (Safe)}
    \label{box_left}
    \includegraphics[width=\textwidth]{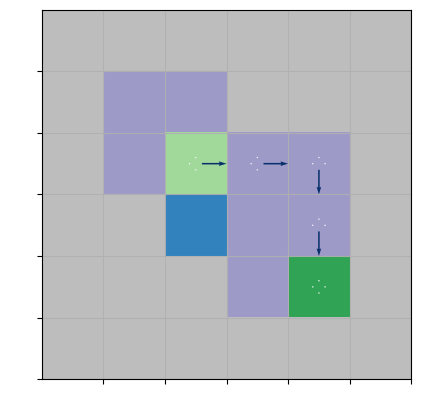}
    \caption[box_down]{Box (Fast)}
    \label{box_down}
  \end{subfigure}
  \caption{Gridworld environments. The initial position is light green, the goal is dark green, the walls are gray, and risky states are purple. Figure (\ref{gridworld}) illustrates 4x4 Marsrover environment. Figure (\ref{box_main})-(\ref{box_down}) illustrates Box environment. In both cases, the agent's task is to get from the initial state to the goal state, and the optimal policy combines with some probabilities fast and safe ways, which are indicated by arrows on the pictures.}
\label{envs}
\end{figure}

To demonstrate the performance of the algorithms, we consider three gridworld environments in our analysis. There are four actions possible in each state, $\mathcal{A} = \{up, down, right, left\}$, which cause the corresponding state transitions, except that actions that would take the agent to the wall leave the state unchanged. Due to the stochastic environment, transitions are stochastic (i.e., even if the agent's action is to go \textit{up}, the environment can send the agent with a small probability \textit{left}). Typically, the gridworld is an episodic task where the agent receives reward -1 on all transitions until the terminal state is reached. We reduce the episodic setting to the infinite-horizon setting by connecting terminal states to the initial state. Since there is no terminal state in the infinite-horizon setting, we call it the goal state instead. Thus, every time the agent reaches the goal, it receives a reward of 0, and every action from the goal state sends the agent to the initial state. We introduce constraints by considering the following specifications of a gridworld environment: Marsrover and Box environments.

\paragraph{Marsrover.} This environment was used in \cite{tessler2018reward, pmlr-v120-zheng20a, NEURIPS2020_Brantley}. The agent must move from the initial position to the goal avoiding risky states. Figure (\ref{gridworld}) illustrates the CMDP structure: the initial position is light green, the goal is dark green, the walls are gray, and risky states are purple. "In the Mars exploration problem, those darker states are the states with a large slope that the agents want to avoid. The constraint we enforce is the upper bound of the per-step probability of step into those states with large slope -- i.e., the more risky or potentially unsafe states to explore" \citep{pmlr-v120-zheng20a}.  Each time the agent appears in a purple state incurs costs 1. Other states incur no cost.

Without safety constraints, the optimal policy is obviously to always go \textit{up} from the initial state. However, with constraints, the optimal policy is a randomized policy that goes \textit{left} and \textit{up} with some probabilities, as illustrated in Figure (\ref{gridworld}). In experiments, we consider two marsorover gridworlds: 4x4 as shown in Figure (\ref{gridworld}) and 8x8.

\paragraph{Box.} Another conceptually different specification of a gridworld is Box environment from \cite{Leike_aisafegrid_2017}. Unlike the Marsrover example, there are no static risky states; instead, there is an obstacle, a box, which is only "pushable" (see Figure (\ref{box_main})). Moving onto the blue tile (the box) pushes the box one tile into the same direction if that tile is empty; otherwise, the move fails as if the tile were a wall. The main idea of Box environment is "to minimize effects unrelated to their main objectives, especially those that are irreversible or difficult to reverse" \citep{Leike_aisafegrid_2017}. If the agent takes the fast way (i.e., goes down from its initial state; see Figure (\ref{box_down})) and pushes the box into the corner, the agent will never get it back, and the initial configuration would be irreversible. In contrast, if the agent chooses the safe way (i.e., approaches the box from the left side), it pushes the box to the reversible state (see Figure (\ref{box_left})). This example illustrates situations of performing the task without breaking a vase in its path, scratching the furniture, bumping into humans, etc.

Each action incurs cost 1 if the box is in a corner (cells adjacent to at least two walls) and no cost otherwise. Similarly to the Marsrover example in Figure (\ref{gridworld}), without safety constraints, the optimal policy is to take the fast way (go down from the initial state). However, with constraints, the optimal policy is a randomized policy that goes down and left from the initial state.

%\begin{figure}
%\centering     %%% not \center
%\subfigure[4x4 gridworld]{\label{fig1:a}\includegraphics[width=0.24\textwidth]{pics/gridworld.png}}
%\subfigure[8x8 marsrover]{\label{fig1:b}\includegraphics[width=0.24\textwidth]{example-image-a}}
%\subfigure[4x4 box]{\label{fig1:b}\includegraphics[width=0.24\textwidth]{pics/box.png}}
%\caption{Environments used in the analysis.}
%\label{regret-plot}
%\end{figure}

\subsection{Simulation results}
\label{subsec:results}

\begin{figure}
    \centering
    \begin{subfigure}{.48\textwidth}
        \includegraphics[width=0.99\textwidth]{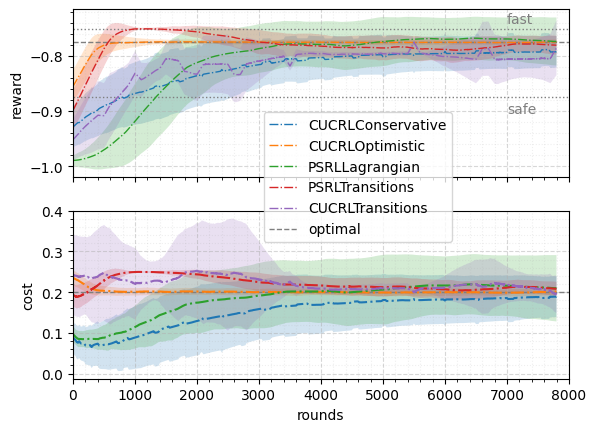}
        \caption[gridworld]{4x4 Marsrover}
        \label{fig2:a}
    \end{subfigure}\hfill
    \begin{subfigure}{.48\textwidth}
        \includegraphics[width=0.99\textwidth]{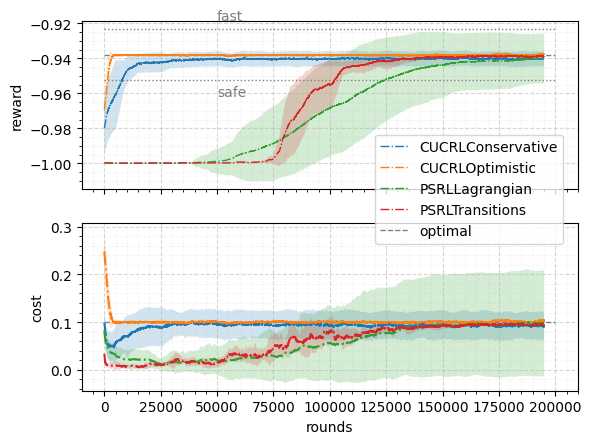}
        \caption[gridworld]{8x8 Marsrover}
        \label{fig2:b}
    \end{subfigure}
    \caption[Caption for Marsrover]%
      {Marsrover results. Two metrics are illustrated: average reward (above) and average cost (below). The dashed line shows the optimal behavior. In reward plots, dotted lines depict the reward level of safe and fast policies. Figure (\ref{fig2:a}) illustrates results for 4x4 Marsrover environment and is averaged over 100 runs with budget 0.2. Figure (\ref{fig2:b}) illustrates results for 8x8 Marsrover environment and is averaged over 30 runs with budget 0.1.}
    \label{marsrover results}
\end{figure}

In this part, we evaluate the performance of the posterior sampling algorithms against the benchmarks. We keep track of two metrics: average reward and average cost. Reward graphs contain three level curves corresponding to the fast, optimal, and safe solutions in each CMDP. Cost graphs have only optimal solution level curves as the safe solution corresponds to 0 budget (and, therefore, 0 cost), and the fast solution corresponds to a certain value, which is high enough to afford the fast way all the time. The source code of the experiments can be found at \url{https://github.com/danilprov/cmdp}.

Figure \ref{marsrover results} compares the average reward and cost incurred by five algorithms for Marsrover environments. \footnote{\textit{CUCRLTransitions} is presented only for 4x4 Marsrover environment. In fact, this algorithm is impractical for even moderate CMDPs because of nonlinear program it solves as a subroutine at the beginning of every epoch (see Appendix \ref{benchmarks} for details).} We observe that all curves converge to the optimal solution in such a simple environment. We can see that \textit{CUCRLOptimistic} gets to optimal solution very fast relatively to the other algorithms. In contrast, it takes longer for \textit{CUCRLConservative}, \textit{PSRLLagrangian},  and \textit{PSRLTransitions} algorithms to get to the optimal solution. In the case of  \textit{CUCRLConservative}, it happens at the expense of safe (conservative) exploration -- the algorithm does not violate constraints during the learning even with random exploration baseline (see Appendix \ref{benchmarks} for more details). In the case of \textit{PSRLLagrangian} and \textit{PSRLTransitions}, this is because of number of parameters that the algorithm learn: while other algorithms estimate $2SA$ number of parameters, posterior sampling algorithms learn $S^2A$ parameters without any knowledge of confidence sets (as Figure (\ref{fig2:b}) shows, an increase from 16 to 64 states slows the learning down considerably).

\begin{figure}
    \centering
    \includegraphics[width=0.7\textwidth]{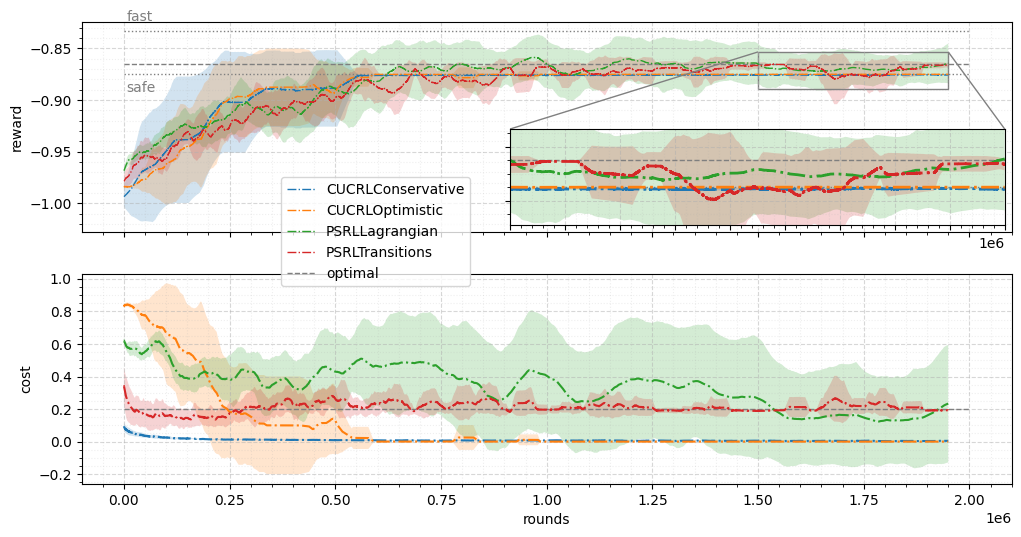}
    \caption[gridworld]{Results for the Box environment; averaged over 30 runs with a budget of 0.2. Two metrics are illustrated: average reward (above) and average cost (below). The dashed line shows the optimal behavior. In reward plots, dotted lines depict the reward level of safe and fast policies.}
    \label{box results}
\end{figure}

Figure \ref{box results} shows results for the Box environment. Both reward and cost graphs show that optimistic algorithms achieve the safe reward value relatively quickly (roughly after 600k rounds). However, those algorithms are stuck with the suboptimal solution afterward, i.e, both algorithms exploit only safe policy ones it is learned. In contrast, it takes a little bit longer for posterior sampling algorithms to deliver a sensible solution, but as the zoomed inset graph shows, both posterior sampling algorithms converge to the optimal solution. Here, we emphasize that
while the absolute difference between algorithms is not significant, the semantics of the achievable values (dotted lines that represent ”safe” and ”fast” solutions) reveal that the existing (OFU-based) algorithms are stuck with a suboptimal solution, while the proposed (posterior-based) algorithms eventually converge to the optimal solution.

Taking a closer look at \textit{PSRLLagrangian} algorithm in Figures \ref{marsrover results}, \ref{box results}, we see that the standard deviation is more expansive than that of \textit{PSRLTransitions}. This can be explained by the fact that for plotting the standard deviation, we used ordinary policy $\pi_k$ instead of mixture policy $\tilde{\pi}_k$ (as mentioned in Section \ref{posterior_algos}). Consequently, the span of the standard deviation highlights that each policy $\pi_k$ is deterministic and may lie far from the optimal solution; however, the mixture policy $\tilde{\pi}_k$ still converges to the optimal solution. As such, the comparison of
proposed algorithms comes down to the fact that \textit{PSRLTransitions} converges faster than \textit{PSRLLagrangian} but at the expense of higher computational costs.

\section{Conclusion}
\label{sec:conclusion}
This paper addresses the practical issue of sample efficient learning in CMDP with infinite-horizon average reward. The experimental evaluation carried out in this paper reveals that posterior sampling is a very effective heuristic for this setting. Compared to feasible optimistic algorithms, we believe that posterior sampling is often more efficient statistically, simpler to implement, and computationally cheaper. In its simplest form, it does not have any parameter to tune and does not require explicit construction of the confidence bound. Consequently, we highlight that the proposed algorithms consistently outperform the existing ones, making them valuable candidates for further research and implementation.

Our work addresses practically relevant RL issues and, therefore, we firmly believe that it may help design algorithms for real-life reinforcement learning applications. The environments presented here expose critical issues of reinforcement learning tasks. Yet, they might overlook problems that arise due to the complexity of more challenging tasks. The following steps, therefore, involve scaling this effort to more complex environments.

Future work also includes a theoretical analysis of the proposed algorithms. As we mentioned, \cite{Agarwal_2021_PSRL} theoretically analyzed a similar posterior sampling algorithm for constrained reinforcement learning. Yet the analysis of \cite{Agarwal_2021_PSRL} is to be revised as our \textit{PSRLTransitions} algorithm is an extension of their algorithm. As regards \textit{PSRLLagrangian}, we anticipate that Lagrangian relaxation would introduce an additional complication in the theoretical analysis and provide some initial thoughts on the analysis of the algorithm in Appendix \ref{apx:theor_alg2}.

% Beyond theoretical aspects, we firmly believe our work may help design algorithms for real-life reinforcement learning applications. 

\begin{ack}
This project is partially financed by the Dutch Research Council (NWO) and the ICAI initiative in collaboration with KPN, the Netherlands.
\end{ack}

\bibliography{main}
\bibliographystyle{plainnat}

\newpage
\appendix

\section{Appendix: theoretical analysis of the Algorithm \ref{alg2:primal_dual_psrl}}
\label{apx:theor_alg2}

\iffalse
\begin{theorem}
 \label{thm:regret_bound}
    The expected regret $\mathbb{E} \left [ Reg_+ (T;r) \right ]$ of Primal Dual Posterior Sampling presented in Algorithm \ref{alg2:primal_dual_psrl} is bounded by
    
    \begin{equation}
        \mathbb{E} \left [ Reg_+ (T;r) \right ] \leq \tilde{O}(\sqrt{T} ).
    \end{equation}
\end{theorem}
\fi

This section provides some insights into the theoretical analysis of the Primal-Dual Posterior Sampling algorithm from Section \ref{posterior_algos}. The complete analysis is left to be covered in the future work.

Let $\hat{r}$ be the optimal solution for the approximate CMDP $\hat{\mathcal{M}} = (\mathcal{S}, \mathcal{A}, \tilde{p}, \tilde{r}, \tau, \rho)$, $\hat{r}_k$ be the solution governed  by policy $\pi_k$ for the approximate relaxed MDP $\hat{\mathcal{M}}_{\lambda}$ (line 7 in Algorithm 1), $r_k$ be the solution governed  by policy $\pi_k$ for the true relaxed MDP $\mathcal{M}_{\lambda}$, and $K_T$ be the number of episodes over horizon $T$.

%{\color{blue} For simplicity consider $Reg(T;r)$, i.e., without $+$ subscript.}

We decompose the reward regret as

\begin{align*}
    Reg (T;r) & = \sum_{t=1}^T \left [ r^* - r(s_t, a_t) \right ] = \sum_{k=1}^{K_T} \sum_{t=t_k}^{t_{k+1}-1} \left [ r^* - r_k \right ] \\
    & = \underbrace{\sum_{k=1}^{K_T} \sum_{t=t_k}^{t_{k+1}-1} \left [ r^* - \hat{r} \right ]}_{(1) \quad \text{PS for CMDP}} + \underbrace{\sum_{k=1}^{K_T} \sum_{t=t_k}^{t_{k+1}-1} \left [ \hat{r} - \hat{r}_k \right ]}_{(2) \quad \text{lagr. relaxation}} + \underbrace{\sum_{k=1}^{K_T} \sum_{t=t_k}^{t_{k+1}-1} \left [ \hat{r}_k - r_k \right ]}_{(3) \quad \text{PS for MDP}}.
\end{align*}

Term (1) denotes how far the transition probabilities and rewards for CMDP $\mathcal{M}$ are from the transition probabilities and rewards induced by the policy at episode $k$. We bound this term by bounding the deviation of sampled probabilities and empirical rewards from the true transition probabilities and rewards. In fact, this is the primary step in the analysis of the Posterior Sampling of Transitions algorithm, which analog is analyzed in \cite{Agarwal_2021_PSRL}.

Term (2) represents the deviation of the approximate CMDP $\hat{\mathcal{M}}$ from the approximate relaxed MDP $\hat{\mathcal{M}}_{\lambda}$. The near-optimality of the relaxation can be proved by leveraging the fact that we are iteratively updating $\pi$ and $\lambda$ using no-regret online learning procedure (Best Response for $\pi$ and
OGD for $\lambda$).

Term (3) brings us back to the true transitions and rewards (where policy $\pi_k$ is actually being executed), showing the deviation between the approximate relaxed MDP $\hat{\mathcal{M}}_{\lambda}$ and the true relaxed MDP $\mathcal{M}_{\lambda}$. We bound this term by the definition of the policy update and the limited maximum span of the value function. Specifically, a similar analysis on expected regret exists \citep{TS_MDP_Ouyang_2017} but under a Bayesian setting. The authors decompose regret into three terms. Two of those terms are replicable in our case, but another heavily relies on the Bayesian property of posterior sampling, which does not hold in our setting. Consequently, the term (3) is where a thorough analysis is needed.

We expect all three terms be bounded by $\tilde{O}( poly(SA) \sqrt{T} )$, where $\tilde{O}$ hides logarithmic factors in $S,A,T$. The cost regret can be bounded following the same reasoning.

\section{Appendix: experimental details}

\subsection{Benchmarks: OFU-based algorithms}
\label{benchmarks}

We borrow three OFU-based algorithms from the existing literature for comparison. These algorithms are based on the first three types of optimistic exploration mentioned in Section 2: double optimism, conservative optimism, and optimism over transitions. Note, we entirely borrow conservative optimism and optimism over transitions type algorithms from the original sources as their setting and assumptions match ours. These algorithms are originally called \textit{C-UCRL} and \textit{UCRL-CMDP} correspondingly. Their names are not informative and might even be confusing. Thus, we let ourselves rename those algorithms as specified above. The double optimism type algorithms was originally developed for the episodic setting, and we borrow only the main principle from it.

All algorithms we consider for empirical comparison solve LP (\ref{eq1})-(\ref{eq4}) as a subroutine substituting unknown parameters $p, r, c_1, ..., c_m$ with their empirical estimates $\hat{p}_t, \hat{r}_{t}, \hat{c}_{1,t}, ..., \hat{c}_{m,t}$. While each algorithm defines the estimates above differently, all of them are based on sample means (\ref{mean_rewards}), (\ref{mean_costs}), and $\Bar{p}_t(s,a,s') = \frac{N_t(s,a,s')}{N_t(s,a) \lor 1}, \quad \forall s, s' \in \mathcal{S}, a \in \mathcal{A}.$

\iffalse
\begin{equation}
    \label{mean_transitions}
    \Bar{p}_t(s,a,s') = \frac{N_t(s,a,s')}{N_t(s,a) \lor 1}, \quad \forall s, s' \in \mathcal{S}, a \in \mathcal{A}.
\end{equation}
\fi

\begin{enumerate}
    \item \textit{CUCRLOptimistic} is a double optimism type algorithm, which implements the principle of optimism under uncertainty by introducing a bonus term $b_t(s,a)$ that favors under-explored actions with respect to each component of reward vector. In the original work, \citet{NEURIPS2020_Brantley} consider an episodic problem; they add a bonus to the empirical rewards (\ref{mean_rewards}) and subtract it from the empirical costs (\ref{mean_costs}):
    
    \begin{align*}
        \hat{r}_t(s,a) =\Bar{r}_t(s,a) + b_t(s,a)
            \quad\mathrm{and}\quad
        \hat{c}_t(s,a) =\Bar{r}_t(s,a) - b_t(s,a).
    \end{align*}
    
    We follow the same principle but recast the problem to the infinite-horizon setting by using the doubling epoch framework described in Section \ref{posterior_algos}.
    
    \item \textit{CUCRLConservative} follows a principle of “optimism in the face of reward uncertainty; pessimism in the face of cost uncertainty.” This algorithm, 
    which was developed by \citet{pmlr-v120-zheng20a}, considers conservative (safe) exploration by overestimating both rewards and costs:
    
    \begin{align*}
        \hat{r}_t(s,a) =\Bar{r}_t(s,a) + b_t(s,a)
            \quad\mathrm{and}\quad
        \hat{c}_t(s,a) =\Bar{r}_t(s,a) + b_t(s,a).
    \end{align*}
    
    \textit{CUCRLConservative} proceeds in epochs of linearly increasing number of rounds $kh$, where $k$ is the episode index and $h$ is the fixed duration given as an input. In each epoch, the random policy \footnote{Original algorithm utilizes a safe baseline during the first $h$ rounds in each epoch, which is assumed to be known. However, to make the comparison as fair as possible, we assume that a random policy is applied instead.} is executed for $h$ steps for additional exploration, and then policy $\pi_k$ is applied for $(k-1)h$ number of steps, making $kh$ the total duration of episode $k$. 
    
    \item \textit{CUCRLTransitions:} Unlike the previous two algorithms, where uncertainty was taken into account by enhancing rewards and costs, 
    \citet{Singh_CMDP_2020} developed an algorithm that constructs confidence set $\mathcal{C}_t$ over $\Bar{p}_t$:
    
    \begin{equation*}
        \mathcal{C}_t = \left \{ p': |p'(s,a,s') - \Bar{p}_t(s,a,s')| \leq b_t(s,a) \quad \forall (s,a) \right \}.
    \end{equation*}
    
    \textit{CUCRLTransitions} algorithm proceeds in epochs of fixed duration of $\ceil*{T^{\alpha}}$, where $\alpha$ is an input of the algorithm. At the beginning of each round, the agent solves the following constrained optimization problem in which the decision variables are (i) Occupation measure $\mu(s,a)$, and (ii) “Candidate” transition $p'$:

    \begin{align}
        \max_{\mu, p' \in \mathcal{C}_t} \sum_{s,a} \mu(s,a) r(s,a), \label{eq1'}\\
        \mathrm{s.t.}\quad \sum_{s,a} \mu(s,a) c_i(s,a) \leq \tau_i,\, \quad i=1,\dots,m, \\
        \sum_a \mu(s,a) = \sum_{s', a} \mu(s', a) p'(s',a,s), \quad \forall s \in \mathcal{S}, \label{eq3'} \\
        \mu(s,a) \geq 0, \quad \forall (s,a) \in \mathcal{S} \times \mathcal{A}, \quad \sum_{s,a} \mu(s,a) = 1, \label{eq4'}
    \end{align}
    
    Note that program (\ref{eq1'})-(\ref{eq4'}) is not linear anymore as $\mu(s', a)$ is being multiplied by $p'(s',a,s)$ in equation (\ref{eq3'}). This is a serious drawback of \textit{CUCRLTransitions} algorithm because, as we show later, program (\ref{eq1'})-(\ref{eq4'}) becomes computationally intractable for even moderate problems.

\end{enumerate}

In all three cases, we use the original bonus terms $b_t(s,a)$ and refer to the corresponding papers for more details regarding the definition of these terms.
\newpage
\subsection{Hyperparameters}
\label{subsec:hyperparam}

\begin{table}[htbp]
  \centering
  \begin{tabular}{ccccccccc}
    \hline
    \multicolumn{2}{c}{Hyperparameter} & \multicolumn{2}{c}{Marsrover 4x4} & \multicolumn{2}{c}{Marsrover 8x8} & \multicolumn{2}{c}{Box 4x4} \\
    \cline{1-8}
    
    \multicolumn{2}{l}{\textit{CUCRLOptimistic}} &  && &&  \\
    \multicolumn{2}{r}{bonus coefficient} &$10^{-2}$&&$10^{-2}$&& 0.5  \\
    %&&&&&  \\
    \cline{1-8}
    
    \multicolumn{2}{l}{\textit{CUCRLConservative}} &   && &&  \\
    \multicolumn{2}{r}{duration $h$} &20&&1000&& 1000  \\
    \multicolumn{2}{r}{bonus coefficient} &$10^{-2}$&&$10^{-2}$&& 0.5  \\
    \cline{1-8}
    
    \multicolumn{2}{l}{\textit{CUCRLTransitions}} &  &&  &&  \\
    \multicolumn{2}{r}{duration $\alpha$} &$1/3$&&-&& -  \\
    \multicolumn{2}{r}{planner $maxiter$} &20&&-&& -  \\
    %&&&&&  \\
    \cline{1-8}
    
    \multicolumn{2}{l}{\textit{PSRLTransitions}} &&  && &&  \\
    \multicolumn{2}{r}{learning rate $\eta$} &$0.2$&&$35*10^{-4}$&& $165*10^{-6}$  \\
    \multicolumn{2}{r}{planner $maxiter$} & 50000&&50000&& 50000   \\
    \multicolumn{2}{r}{planner $tolerance$} & $10^{-3}$ && $10^{-3}$&& $10^{-3}$ \\
    \multicolumn{2}{r}{planner $\gamma$} & 0.95 && 0.95&& 0.95 \\
    \cline{1-8}
    
    \multicolumn{2}{l}{\textit{PSRLLagrangian}} &  &&  && \\
    \multicolumn{2}{r}{-} &\multicolumn{6}{r}{does not require any hyperparameters}  \\
    %&&&&&  \\
    \cline{1-8}

    \hline
   
  \end{tabular}
  \caption{Selected Hyperparameters}
  \label{tbl:hyperparameters}
\end{table}

\end{document}